\documentclass[10pt,twocolumn,letterpaper]{article}

\usepackage{cvpr}              

\usepackage{graphicx}
\usepackage{subcaption}

\definecolor{cvprblue}{rgb}{0.21,0.49,0.74}
\usepackage[pagebackref,breaklinks,colorlinks,allcolors=cvprblue]{hyperref}

\def\paperID{00000} 
\def\confName{CVPR}
\def\confYear{2026}

\title{CAD 100K: A Comprehensive Multi-Task Dataset for Car Related Visual Anomaly Detection

 }

\author{Jiahua Pang \quad Ying Li \thanks{Corresponding author.}  \quad  Jingcai Luo\quad Yanuo Zheng \quad Bao Yunfan \quad Yujie Lei  \quad Yuxi Tian \\
Beijing Institute of Technology\\
\and 
Dongpu Cao \thanks{Corresponding author.} \
Tsinghua University\\
\and
Rui Yuan\
China Agricultural University \\
\and 
Guojin Yuan\
Beijing Jiaotong University\\
\and
Hongchang Chen\
The Hong Kong Polytechnic University\\
\and
Zhi Zheng \quad Yongchun Liu\
Li Auto\\
}
\usepackage{amsmath}
\usepackage{amssymb}
\usepackage{booktabs}
\usepackage{afterpage}
\usepackage{multirow}
\def\confName{CVPR}
\begin{document}
\maketitle
\begin{figure*}[t]
    \centering
    \includegraphics[width=1\linewidth]{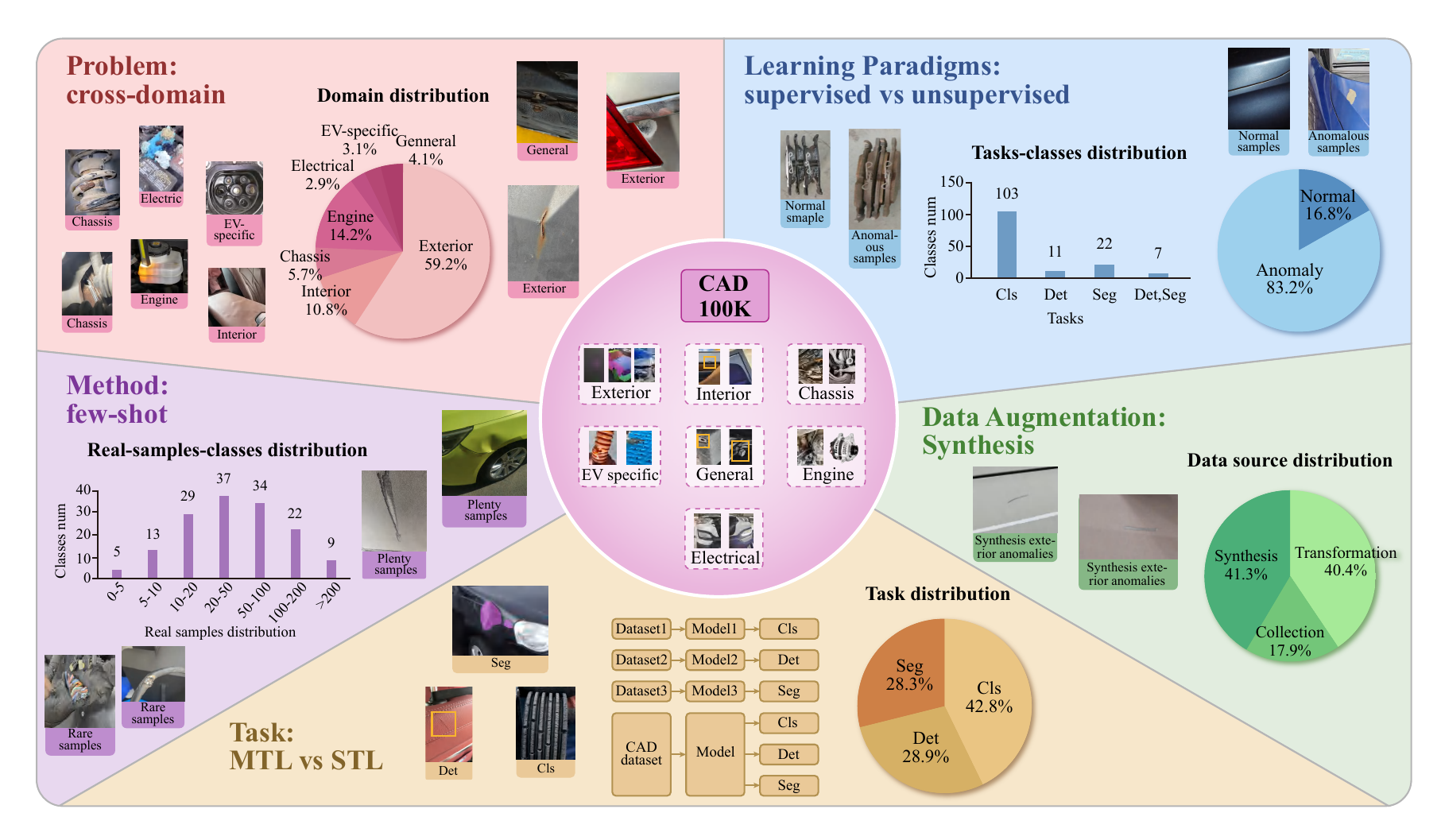}
    \caption{\textbf{Overview of the CAD 100K Dataset.} The CAD 100K dataset contains over 100$K$ images crossing 7 domains and 3 tasks, providing a comprehensive view for car-related anomaly detection.  It features a multi-task architecture and synthesis data augmentation methods, solves cross-domain problems, supports both supervised and unsupervised learning paradigms, and is compatible with few-shot learning frameworks.}
    \label{fig:example}
  \end{figure*}
\begin{abstract}

Multi-task visual anomaly detection is critical for car-related manufacturing quality assessment. However, existing methods remain task-specific, hindered by the absence of a unified benchmark for multi-task evaluation. To fill in this gap, We present the CAD Dataset, a large-scale and comprehensive benchmark designed for car-related multi-task visual anomaly detection. The dataset contains over 100$K$ images crossing 7 vehicle domains and 3 tasks, providing models a comprehensive view for car-related anomaly detection. It is the first car-related anomaly dataset specialized for multi-task learning(MTL), while combining synthesis data augmentation for few-shot anomaly images. We implement a multi-task baseline and conduct extensive empirical studies. Results show MTL promotes task interaction and knowledge transfer, while also exposing challenging conflicts between tasks. The CAD dataset serves as a standardized platform to drive future advances in car-related multi-task visual anomaly detection.

\end{abstract}    
\section{Introduction}
\label{sec:intro}
Vision-based anomaly detection using multi-task learning (MTL) scheme is a core technology \cite{graham2023one,baitieva2024supervised,zhang2024pku,mvtec_ad,mallios2023vehicle,hasan2025vehicle} for smart manufacturing in the automotive industry, which simulates human visual capabilities through computer vision to achieve precise identification, localization, and classification of product defects. This task covers from the overall vehicle appearance to micro defects, ensuring the reliability and safety of the entire process from automotive components to vehicle integration in a comprehensive manner \cite{peng2025car,wang2023cardd}. However, existing anomaly detection methods \cite{baitieva2024supervised,schmidl2022anomaly,deng2022anomaly} mainly employ single-task learning (STL) models to detect one or a few types of anomalies. Although these models can achieve high performance with a low computation cost, they struggle to meet the diverse anomaly detection requirements of automotive manufacturing, leading to a proliferation of specialized STL models and a high reliance on data. Existing relevant datasets \cite{wang2023cardd,huynh2023vehide,peng2025car,zhang2024pku,mvtec_ad,real_ad} are also designed for these STL models, lacking a unified multi-task dataset to cover vehicle-related anomaly detection tasks. 

To fill this gap, we present CAD 100$K$, a large-scale and comprehensive benchmark  designed for car related multi-task visual anomaly detection. Our proposed dataset comprises more than 100,000 images spanning seven vehicle domains and three distinct tasks. This dataset is collected via three methods: real-world data collection, open-source dataset conversion, and synthetic data generation. Following rigorous data cleaning, all data is structured in a domain–system–part hierarchy and annotated for industrial downstream application. It features a multi-task architecture, supports both supervised and unsupervised learning paradigms, and is compatible with few-shot learning frameworks. 

We further introduce a multi-task baseline framework specifically designed for our CAD 100K dataset. Our baseline adopts a shared-backbone multi-head structure, supporting both CNN-based and ViT-based encoders. A convergence-aware balancing mechanism is introduced to dynamically adjust inter-task weights during joint optimization. We employ synchronized mixed-precision training across all heads, with shared early layers frozen for stability during warm-up. Additionally, we conduct comprehensive experiments to comparatively evaluate the performance of advanced single-task models against our proposed multi-task baseline. Our key contributions are outlined as follows:
\begin{enumerate}
    \item We introduce CAD 100K, a large-scale vision anomaly detection dataset comprising 100,000+ anomaly images spanning seven vehicle domains and three distinct tasks.
    \item We propose the first multi-task baseline model tailored for car-related industrial anomaly detection, which simultaneously learns from shared visual information across multiple domains.
    \item We benchmark public datasets to analyze the performance of advanced single-task models against our multi-task baseline, and validate our approach on the CAD 100K dataset to demonstrate our collected data quality.
\end{enumerate}

\section{Related Work}
\subsection{Car-related Anomaly Detection Datasets}

Car-related anomaly detection has evolved from early classification-focused datasets to more specialized benchmarks \cite{wang2023cardd,susutti2024real,peng2025car}. For example, CarDD \cite{wang2023cardd} provides 4,000 high-resolution images with over 9,000 annotated instances across six damage types, supporting classification, object detection and instance segmentation. The Car Parts and Damages Dataset \cite{susutti2024real} comprises 1,812 images (998 for car parts, 814 for damages) and 24,851 polygon annotations for fine-grained part localization and damage segmentation. 

However, these datasets remain limited in two key aspects. First, they are mainly designed for single tasks, whereas real-world anomaly detection requires multi-task integration—including classification, part identification, segmentation, and severity assessment. Second, the majority of existing car-related anomaly datasets focus predominantly on obvious exterior damage \eg, glass shatter, lamp broken) and overlook internal mechanical faults, structural deformations or complex environmental anomalies.

Merging existing datasets introduces multiple challenges: inconsistent annotations (e.g., differing label sets and segmentation protocols), conflicting labels (e.g., what constitutes a “scratch” vs “dent”), and varying part/detail granularity \cite{hernandez1998real}. This is exacerbated by a pronounced domain gap, stemming from disparities in image resolution, viewpoint, anomalous type distribution and scene context, as well as an imbalance in task difficulty (e.g., segmentation vs classification) \cite{oquab2014learning}. Together, these factors create conflicting learning signals that hinder model generalization and robustness.

Thus, a comprehensive, multi-task benchmark, covering diverse anomaly types and scenarios and designed for task compatibility and scalability, is essential to advance real-world car-related anomaly detection.
\subsection{Multi-Task Learning anomaly detection related methods}

Multi-task learning (MTL) is suitable for the task of car-related anomaly detection, and has been widely applied in other domains like LLM, remote sensing. Li et al.  \cite{li2024co} introduce the idea of applying one shared-backbone frameworks for different types of tasks including classification, object detection and semantic segmentation.

Although Li et al. \cite{li2024co} achieve better performance than single-task learning tasks in specific remote-sensing task, there are several challenges when applying this idea to car-related anomaly detection tasks.  Firstly, a foundational difficulty arises from heterogeneous data sources: the image resolution varies in different datasets, or even in one dataset itself, which limits the usage of transfromer backbones \cite{cai2023itran}. Furthermore, task-difficulty imbalance makes it harder for MTL-learing to surpass STL-learning method in every task \cite{yun2023achievement}. One remedy in the literature is dynamic loss-weighting (\eg, methods such as Conflict-Aware Balance, CoBa \cite{yang2025cabs}) which adapts the relative importance of tasks via loss-level weights. However, such weighting schemes operate at the loss level and cannot resolve deeper issues of gradient conflict during back-propagation, where shared parameters receive conflicting gradient signals from multiple tasks. Methods like GradNorm \cite{chen2018gradnorm} aim to balance task training by manipulating gradient magnitudes, thereby mitigating conflict and improving overall Pareto performance. Nonetheless, its performance gain on particularly difficult tasks can be limited.

In conclusion, a robust MTL framework for complex domains like car-related anomaly detection likely requires a synergistic approach that co-designs adaptive architectures, dynamic optimization strategies, and sophisticated gradient coordination \cite{wei2021review}.



\section{CAD 100K Dataset}
\label{sec:cad_dataset}
\begin{figure*}[t]
    \centering
    \includegraphics[width=1\linewidth]{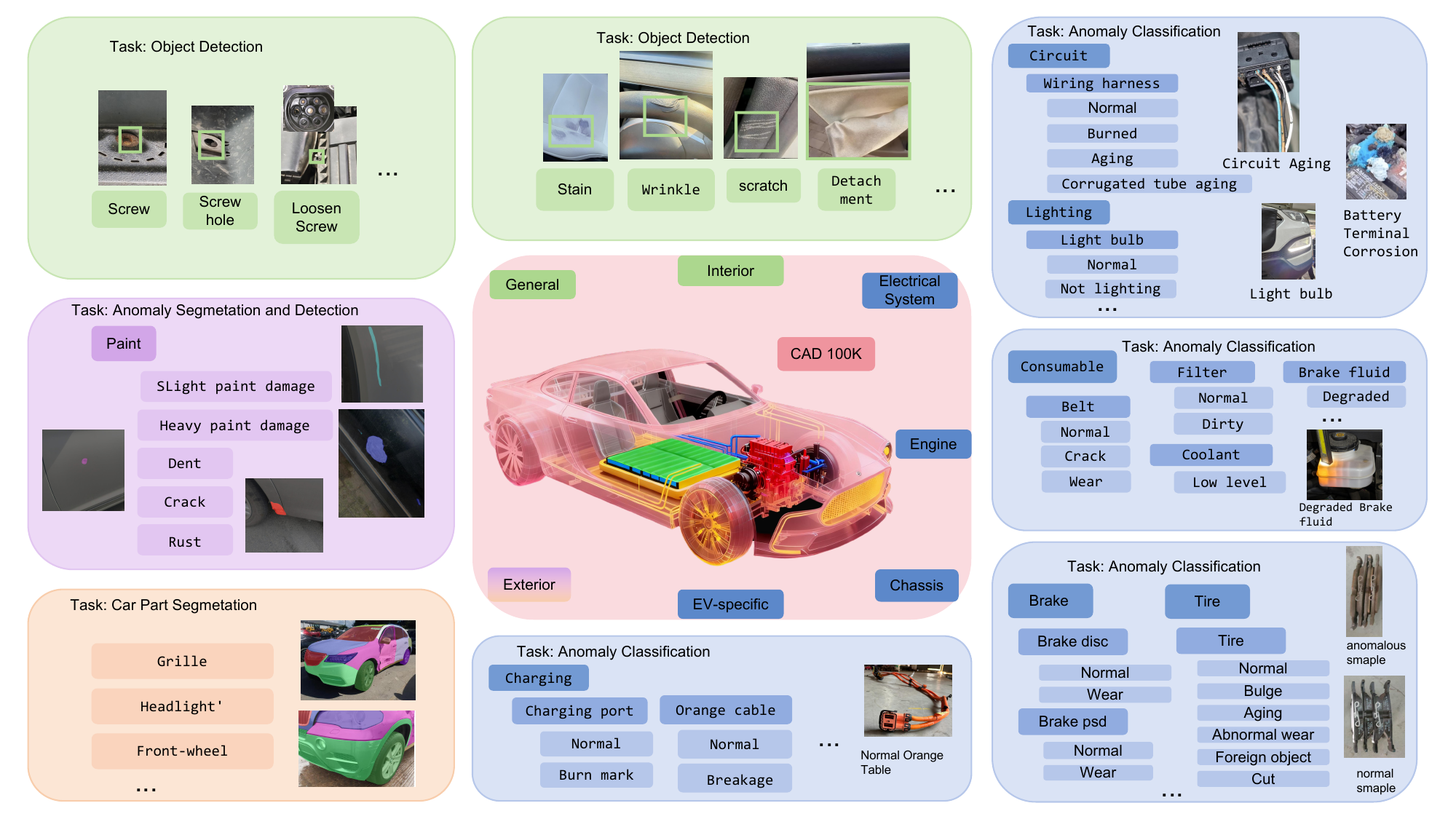}
    \caption{\textbf{Tasks and structure of the CAD 100K Dataset.} CAD 100K adopts a hierarchical domain–system–part anomaly structure to ensure semantic consistency and extensibility across different tasks.  }
    \label{fig:structure}
  \end{figure*}
The CAD 100K Dataset provides a unified, large-scale benchmark for \textit{multi-task car-related anomaly detection}, addressing the lack of a standardized dataset that jointly supports classification, detection, and segmentation under real-world automotive conditions. 
It establishes a hierarchical organization across domains and systems, integrates real and synthetic data, and is designed to enable analysis of robustness, transferability, and task interaction within multi-task learning (MTL) frameworks.
\subsection{High-Level Goals}
\label{sec:cad_goals}

\textbf{Comprehensive anomaly coverage.} To comprehensively cover visual anomaly detection tasks related to vehicles, we systematically divide the detection areas based on multiple categories including vehicle exterior, chassis, cabin interior, and components. Corresponding data collection and organization are then conducted for each specific detection domain.

\textbf{Multi-Task specialization.} The downstream application of industrial anomaly detection typically involve multiple types of tasks such as detection, segmentation, and classification \cite{yan2024comprehensive}, which may be executed either simultaneously or separately within the same workflow . Our constructed dataset and baseline can meet these practical needs in industrial scenarios.

\textbf{Support for multiple learning paradigms.} With multi-level anomaly annotations (classification, detection, segmentation) and abundant normal samples, the CAD 100K dataset supports both fully-supervised and unsupervised anomaly detection paradigms \cite{omar2013machine}.

\textbf{Compatibility with few-Shot learning.} The core objective of industrial production lines is to ensure high yield rates, which inherently results in a very limited number of anomaly samples available \cite{mai2006sampled}. Consequently, many anomaly detection tasks \cite{huyan2022aud,wei2024few,fang2023fastrecon} need to address the challenge of few-shot learning \cite{chen2024survey}. In response, this dataset has specifically designed few-shot categories to adapt to and facilitate efficient few-shot learning approaches.

\textbf{Open-set friendly.} Our dataset is structured in a hierarchical \textit{domain–system–part} taxonomy. For unseen anomaly data types, new categories can be added based on the dataset structure.


\subsection{Dataset Structure and Design Principles}
\label{sec:cad_structure}

Unlike prior single-task datasets, CAD 100K adopts a hierarchical \textbf{domain–system–part} anomaly structure to ensure semantic consistency and extensibility across different tasks. The CAD 100K dataset spans 7 domains, 23 systems, and over 78 anomaly classes, providing extensive coverage of car-related defects across both appearance and functional components.
As illustrated in Fig.~\ref{fig:structure}, it is organized along domains, systems and parts, and tasks.

 \textbf{Domains (7 domains).} This dataset includes \textit{Exterior}, \textit{Interior}, \textit{Chassis}, \textit{Engine Bay}, \textit{Electrical}, \textit{EV-specific}, and \textit{General} domians.  
Domains are defined based on subsystem anomaly statistics, visual detectability, and interface for new anomalous.  
The \textit{General} domain encompasses cross-system anomalies (\eg bolt loosening) and ensures future extensibility.
  
\textbf{Systems and Parts (23 systems, 78 parts).} Each domain is decomposed into functional subsystems (\eg suspension, lighting, charging) and associated parts (\eg wheel, lamp, battery pack), capturing both structural and functional hierarchy for fine-grained anomaly localization.
  
\textbf{Tasks (3 types and variants).} In our dataset, all data is partitioned and annotated based on actual downstream application needs, including tasks such as classification, segmentation, and detection. For example, classification tasks include part and damage-type recognition and normal/anomaly discrimination; detection tasks include localization of defective regions or missing components; segmentation tasks include pixel-level parsing of part boundaries or surface damages.

Each sample follows a unified semantic linkage:
\[
\text{domain} \rightarrow \text{system} \rightarrow \text{part} \rightarrow \text{anomaly type} \rightarrow \text{tasks},
\]
ensuring cross-task consistency and scalable annotation.

The domain and system taxonomy in CAD 100K is defined by three practical dimensions:

\begin{itemize}
    \item System structure and anomaly frequency: CAD 100K’s organization mirrors the physical vehicle structure—domains, systems and parts align with real car construction. In fact, the anomaly rates vary across different domains. Since exterior components are designed with a focus on aesthetics and lightweighting, and are highly exposed to external impacts \cite{berwo2024vebd}, their  anomaly rate tends to be higher than that of the cabin and other interior parts. By selecting parts and domains with these patterns in mind, CAD 100k ensures efficient anomaly acquisition and includes both frequent and rare failure modes.
    \item Visual detectability: Vehicle subsystems differ markedly in visual clarity and annotation complexity. Exterior panels support fine‐grained segmentation under uniform lighting; whereas interior zones or under-body regions are visually cluttered and less suited to pixel-level masks \cite{mallios2023vehicle}. Recognizing these constraints, CAD 100k assigns segmentation/detection to high visibility domains, and classification to lower visibility ones.
    \item Interface for new anomalous: To remain forward-compatible, CAD 100K introduces an “EV-specific” domain (battery pack, drive motor, charging port) and a “General” domain for cross-system faults (e.g., fastener loosening). This ensures the dataset’s relevance to both current and future vehicle architectures.
\end{itemize}

Together, these principles align CAD 100K with industrial reliability data, perceptual feasibility, and MTL requirements.

\begin{figure*}[t]
    \centering
    \includegraphics[width=1\linewidth]{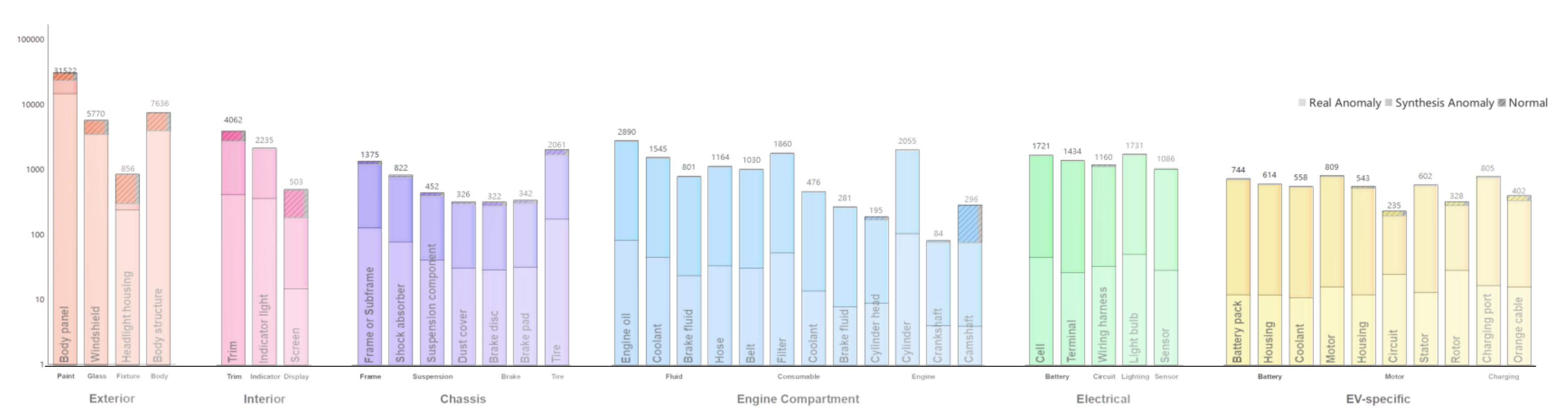}
    \caption{\textbf{The data composition across domains, tasks, and sources in CAD 100K.}  All images are in RGB format with various resolutions from 100 × 84 to 4096 × 3072, covering diverse lighting and viewpoints.}
    \label{fig:statistics}
  \end{figure*}

\subsection{Dataset Acquisition and Processing}
\label{sec:cad_construction}

 CAD 100K is built through a systematic pipeline that combines real-world acquisition with synthetic generation, ensuring both broad coverage of automotive anomalies and consistent hierarchical semantics.
 
\textbf{Real-world data collection.}  
We deploy an image capture system across multiple automotive service center and assembly facilities to collect high-resolution photographs of car components under various viewpoints, lighting conditions and damage states. Each image is annotated according to our domain–system–part–anomaly taxonomy with bounding boxes for detection and pixel-masks for segmentation.

\textbf{Open-source transformation.}  
To expand dataset diversity, we incorporate existing car damage image datasets \cite{wang2023cardd,lin2022augmentation,susutti2024real} by  re-mapping their labels into our unified hierarchy,  filtering low-quality or ambiguous samples, and refining annotations to align with our task schema (classification, detection, segmentation).

\textbf{Synthetic data generation.}  
To address long-tail rarity of certain defect types, we designed a multi-stage synthetic pipeline:  
 Based on generative diffusion models \cite{dai2025seas,song2023objectstitch}, we synthesize plausible defects (e.g., corrosion, surface crack, missing element) onto those templates. Anomalous templates are composited into real backgrounds with realistic lighting and texture transformations, and matching “normal” variants are produced for classification and unsupervised tasks.  This approach improves not only sample volume but critical coverage of hard-examples (e.g., low-contrast, occluded or subtle damages), thus improving evaluation of model robustness and synthesis-to-real transfer.

\textbf{Hierarchical classification and organization.}  
All samples (real or synthetic) are catalogued using the domain–system–part anomaly task linkage, enabling  fine-grained performance analysis, controlled sampling for class and task balance, and efficient retrieval for specific experimental configurations.

\textbf{Data cleaning and validation.}  
We apply automated and manual quality control: removing corrupted or misaligned images, validating annotations via multi-expert review, cross-checking annotation consistency across tasks (classification labels, detection boxes, segmentation masks), and performing statistical audits to identify annotation bias.
In the final release, the dataset comprises approximately 53\% real-world images and 47\% synthetic ones—optimally balanced to maximise both realism and anomaly-type diversity while respecting the hierarchical structure necessary for rigorous multi-task evaluation.
\subsection{Dataset Statistics and Methodological Analysis}
\label{sec:cad_statistics}
Figure~\ref{fig:statistics} summarizes data composition across domains, tasks, and sources.  
All images are in RGB format with resolutions from $100\times84$ to $4096\times3072$, covering diverse lighting and viewpoints.

\textbf{Domain imbalance.}
The \textit{Exterior} domain contributes about 60\% of all samples, consistent with real-world visibility and the prevalence of appearance-level defects.  
Domains such as \textit{Engine Bay}, \textit{Interior}, and \textit{Electrical} are relatively underrepresented due to acquisition difficulty, forming a realistic cross-domain imbalance crucial for domain adaptation research.

\textbf{Task distribution.}
Task ratios follow domain visibility: exterior scenes mainly support segmentation/detection, while interior and mechanical scenes favor classification.  
This coupling naturally supports research on task synergy and interference within MTL frameworks.

\textbf{Synthetic expansion.}
To enrich anomaly diversity and balance distributions in type, color, and spatial position, the CAD 100K dataset incorporates synthesized images generated by both general-purpose and industry-specific models. Diffusion-based synthesis is used here to enlarge the dataset. More importantly, synthesis targets \textit{hard examples},like complex paint damages including multi-types, dents without clear edges, enhancing coverage of visually challenging cases and enabling rigorous evaluation of hard-case robustness and synthetic-to-real transfer.

\textbf{Few-shot and long-tail subsets.}
Long-tail distributions persist across categories, with many part-level anomalies having fewer than 50 instances.  
These subsets facilitate few-shot and data-efficient anomaly detection, where models must generalize from head classes or auxiliary domains.

\textbf{Supervision paradigms.}
Normal samples comprise 20\% of CAD 100K, enabling both fully supervised and hybrid semi-unsupervised training.  
The dataset thus bridges dense anomaly annotation with representation learning from normal data, supporting experiments across different supervision regimes.

In general, CAD 100K integrates domain diversity, synthetic enhancement, and hierarchical task alignment into a coherent MTL-oriented benchmark, bridging dataset construction and methodological investigation in car anomaly detection.

 \section{Baseline}
\label{sec:baseline}

To efficiently address the diverse objectives of car-related anomaly detection—classification, detection, and segmentation, we design a unified MTL-oriented baseline that harmonizes shared feature learning across tasks, while adapting dynamically to their varying convergence speeds and data complexity. Our baseline adopts a shared-backbone multi-head structure inspired by \textit{RSCoTr} \cite{li2024co}, supporting both CNN- and ViT-based encoders. 
\begin{figure}[t]
    \centering
    \includegraphics[width=1\linewidth]{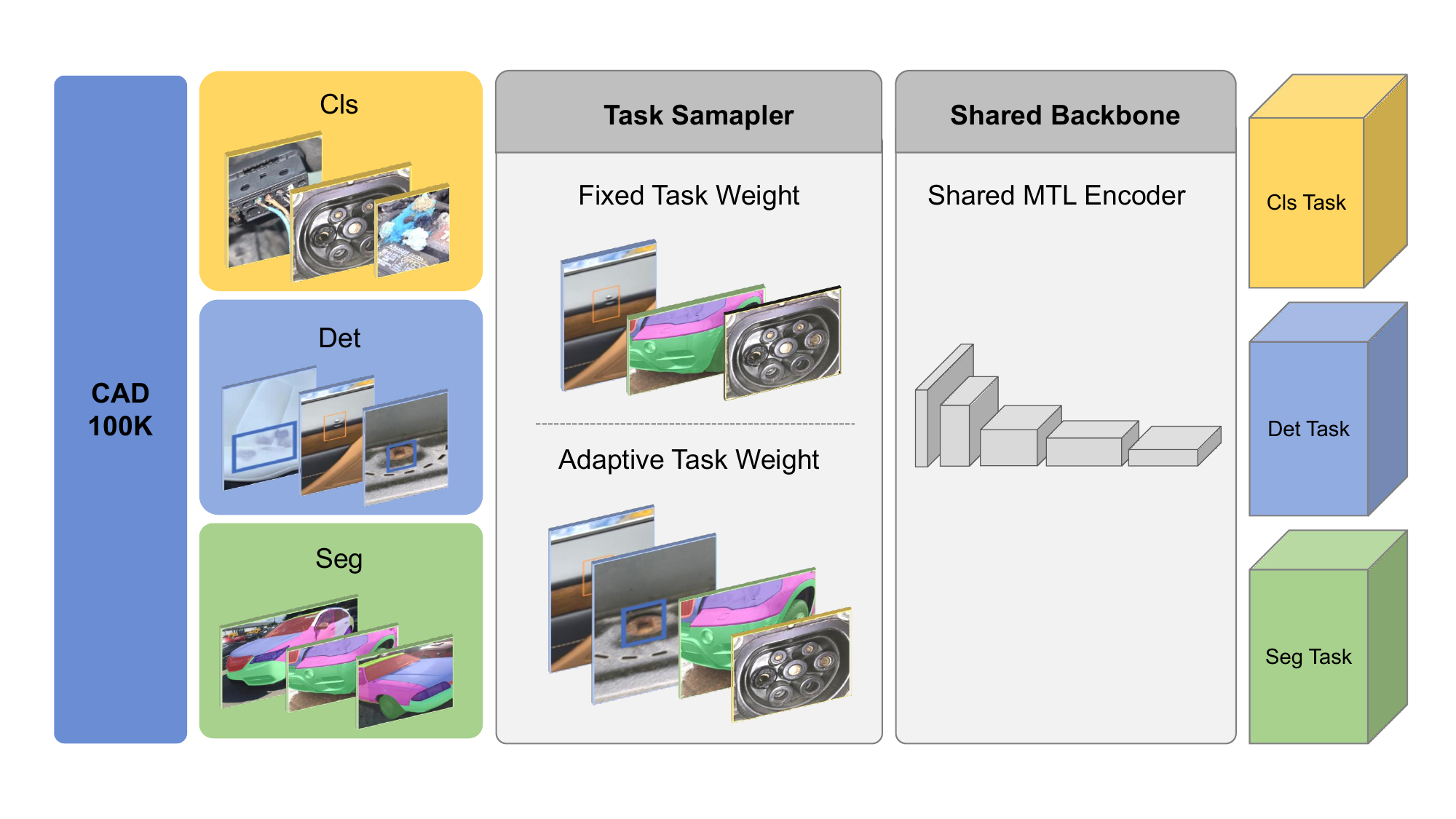}
    \caption{\textbf{Structure of the proposed Baseline.}    }
    \label{fig:example}
  \end{figure}

Specifically, we implement a shared backbone across all domains, ensuring unified feature extraction for classification, detection and segmentation tasks. The choice between ConvNeXt \cite{liu2022convnet} and DINOv3 \cite{simeoni2025dinov3} is determined by deployment constraints and model scaling rather than domain separation—both serve as a common visual encoder for the entire benchmark.
The architecture maintains a unified input interface across tasks of different resolutions, with lightweight task-specific decoders (classification, detection, segmentation) attached atop shared representations.
This structure enables consistent feature sharing and efficient adaptation to scene-dependent resolution variation.

\textbf{Adaptive Task Co-Training.} We introduce a balancing mechanism considering convergence-CoBa (Convergence Balancer)-to dynamically adjust inter-task weights during joint optcimization. 
CoBa\cite{gong2024coba} evaluates each task’s relative convergence speed via the rate of loss descent and stability over a moving window:
\begin{equation}
w_t \propto \alpha\, \text{RCS}_t + \beta\, \text{ACS}_t + \gamma\, \text{DF}_t,
\label{eq:1}
\end{equation}
where RCS, ACS, and DF denote relative, absolute, and divergence-based convergence scores respectively. Tasks with slower or unstable convergence receive higher weights, ensuring balanced learning progress and preventing overfitting of dominant tasks.

\textbf{Task Scheduler and Data Sampling.} On top of CoBa\cite{gong2024coba}, we design an adaptive task scheduler that selects which task to train at each iteration according to softmax-normalized CoBa\cite{gong2024coba} priorities. 
Tasks with higher weights (\ie, poorer convergence) are sampled more frequently. 
The associated data loader synchronizes sampling probabilities accordingly:
\begin{equation}
P(\text{task}_i) = \frac{\exp(w_i / T)}{\sum_j \exp(w_j / T)},
\label{eq:2}
\end{equation}
where $T$ is the temperature controlling stochasticity. 
This cooperative mechanism adaptively allocates computational focus toward under-trained tasks while maintaining global convergence stability.

\textbf{Training and Optimization.} We employ synchronized mixed-precision training across all heads, with shared early layers frozen for stability during warm-up. 
Losses are normalized per task and dynamically reweighted using CoBa\cite{gong2024coba} outputs. 
This adaptive balancing yields faster convergence and improved generalization in challenging multi-domain scenarios, outperforming static-weight MTL baselines in both anomaly localization and part classification.

Overall, the proposed baseline serves as a strong reference for future research in multi-task car anomaly detection, demonstrating how convergence-driven coordination can unify task learning across heterogeneous visual conditions.

\section{Experiments}

\subsection{Experimental Settings}

\paragraph{Datasets and Evaluation Protocols.} To evaluate the effectiveness of our proposed MTL-oriented baseline comprehensively, we conduct rigorous experiments across two distinct settings. We perform extensive benchmarking on publicly available datasets to systematically analyze the performance characteristics of different backbone architectures under both Single-Task Learning (STL) and Multi-Task Learning (MTL) paradigms. Besides, we validate our approach using the real-world collected data in our CAD 100K dataset to demonstrate its practical utility in industrial applications.

For public benchmarks, we carefully select three complementary datasets that collectively cover the spectrum of visual understanding tasks in automotive anomaly detection:

\begin{itemize}[leftmargin=*,nosep]
    \item 
    Car Part 50 Classification Dataset \cite{lin2022augmentation}: Comprising 50 fine-grained car part categories, focusing on interior and exterior component recognition.
    \item 
    CarDD Detection Dataset \cite{wang2023cardd}: Containing bounding box annotations for various car damages and defects, including scratches, dents, and cracks.
    \item 
    Car Parts Segmentation Dataset \cite{susutti2024real}: Providing pixel-level annotations for car part segmentation from the Car Parts and Damages dataset.
\end{itemize}

These datasets span classification, detection, and segmentation tasks, enabling us to compare STL versus shared MTL. Notably, the semantic overlap among these tasks is limited, making this setting particularly challenging and primarily assessing our baseline's capability to generalize across weakly related tasks.
\paragraph{Evaluation Metrics.}
We employ task-specific evaluation metrics following established practices in each domain:

\begin{itemize}[leftmargin=*,nosep]
    \item Classification: We report Overall Accuracy (OA) and Top-1 Accuracy to measure categorical recognition performance.
    
    \item Segmentation: We utilize mean Intersection-over-Union (mIoU) and mean F1 Score (mF1). The F1 score is computed as the harmonic mean of precision and recall for each class, then averaged across categories: 
    \begin{equation}
    \text{mF1} = \frac{1}{C}\sum_{c=1}^{C} \frac{2 \cdot \text{Precision}_c \cdot \text{Recall}_c}{\text{Precision}_c + \text{Recall}_c},
    \label{eq:3}
    \end{equation}
    where $C$ is the number of classes.
    
    \item Detection: We adopt the standard MS COCO evaluation protocol, reporting mean Average Precision (mAP) averaged over IoU thresholds from 0.50 to 0.95.
\end{itemize}

\paragraph{Implementation Details.}
Our implementation builds upon PyTorch with careful attention to reproducibility. We employ multiple backbone variants to study scale effects: DINOv3-Splus (smallsplus), DINOv3-B (base), and DINOv3-L (large)\cite{simeoni2025dinov3}. All models are trained from pre-trained weights using AdamW optimizer with initial learning rate of $1\times10^{-4}$ and cosine annealing schedule. We use batch size of 32 per GPU and train for 100 epochs.

We compare three distinct training paradigms:
\begin{itemize}[leftmargin=*,nosep]
    \item Single-Task (STL): Individual models trained separately for each task
    \item Base MTL: Shared backbone with round-robin task scheduling and equal loss weighting
    \item Adaptive MTL (Ours): Our proposed method with CoBa dynamic weighting and adaptive task scheduling
\end{itemize}

\subsection{Public Benchmark Evaluation}
\begin{table}[t]
\centering
\caption{Comprehensive performance comparison on public car-related benchmarks across different backbone architectures, scales, and training paradigms. Results show classification accuracy (Cls. Acc), detection mAP (Det. mAP), and segmentation mIoU (Seg. mIoU). Bold indicates the best performance within each backbone size and architecture combination.}
\resizebox{0.48\textwidth}{!}{
\begin{tabular}{l|ccccc}
\toprule
\textbf{Backbone Type} & \textbf{Backbone Size} & \textbf{Method} & \textbf{Cls. Acc (\%)} & \textbf{Det. mAP (\%)} & \textbf{Seg. mIoU (\%)} \\
\midrule
\multirow{9}{*}{DINOv3 (ViT)}
& \multirow{3}{*}{Smallplus} 
& Single & \textbf{98.40} & \textbf{59.20} & \textbf{72.80} \\
& & Base & \textbf{98.40} & 58.90 & 70.93 \\
& & Adaptive & \textbf{98.40} & 57.80 & 71.88 \\
\cmidrule{2-6}
& \multirow{3}{*}{Base}
& Single & \textbf{100.00} & \textbf{59.40} & \textbf{72.42} \\
& & Base & 98.80 & 58.20 & 71.95 \\
& & Adaptive & 99.10 & 58.90 & 72.25 \\
\cmidrule{2-6}
& \multirow{3}{*}{Large}
& Single & \textbf{100.00} & \textbf{62.10} & \textbf{73.53} \\
& & Base & 99.20 & 61.80 & 72.65 \\
& & Adaptive & 99.20 & 61.70 & 72.40 \\
\midrule
\multirow{9}{*}{ConvNeXt (CNN)}
& \multirow{3}{*}{Small}
& Single & \textbf{98.80} & 58.80 & \textbf{72.80} \\
& & Base & \textbf{98.80} & 58.60 & 71.12 \\
& & Adaptive & \textbf{98.80} & \textbf{59.10} & 71.82 \\
\cmidrule{2-6}
& \multirow{3}{*}{Base}
& Single & 98.60 & \textbf{60.40} & \textbf{72.8} \\
& & Base & 98.80 & 60.10 & 71.95  \\
& & Adaptive & \textbf{99.20}& 59.60  &71.31  \\
\cmidrule{2-6}
& \multirow{3}{*}{Large}
& Single & \textbf{100.00} & \textbf{62.10} & \textbf{73.38} \\
& & Base & \textbf{100.00} & 60.70 & 72.58\\
& & Adaptive & 98.90 & 61.70 & 71.90 \\
\bottomrule
\end{tabular}
}
\label{tab:public_results_combined}
\end{table}
\paragraph{Comparative Analysis.}
\paragraph{Comparative Analysis}
Table~\ref{tab:public_results_combined} presents comprehensive results across different backbone architectures, scales, and training paradigms. The experimental findings reveal several noteworthy patterns:

\textbf{Performance Saturation in Classification.} The classification task demonstrates remarkable performance saturation across all configurations, with multiple architectures achieving perfect 100\% accuracy in single-task settings (DINOv3-Base, DINOv3-Large, and ConvNeXt-Large). This indicates that car part classification represents a relatively solved problem even with moderate model capacities, creating significant optimization challenges for multi-task learning where gradient dominance from simpler tasks may impede learning in more complex ones.

\textbf{Multi-task Learning Dynamics.} The fixed task-weight MTL (Base scheme) exhibits complex behavior across different architectures. For DINOv3-Smallplus, base MTL maintains classification accuracy (98.4\%) while experiencing modest degradation in detection (59.20\%→58.90\%) and more significant segmentation decline (72.80\%→70.93\%). Interestingly, ConvNeXt-Small demonstrates different characteristics, with base MTL preserving classification performance (98.8\%) while showing minimal detection degradation (58.80\%→58.60\%) and moderate segmentation decline (72.80\%→71.12\%).

\textbf{Adaptive Strategy Performance.} Our adaptive MTL approach exhibits architecture-dependent effectiveness. In DINOv3-Smallplus, adaptive MTL shows mixed results—improving segmentation performance (70.93\%→71.88\%) but experiencing detection degradation (58.90\%→57.80\%). Conversely, ConvNeXt-Small demonstrates the adaptive strategy's potential, achieving the best detection performance (59.10\%) while maintaining classification accuracy (98.8\%) and improving segmentation (71.12\%→71.82\%) over base MTL.

\textbf{Backbone Scale and Architecture Effects.} Scaling effects vary significantly between architectures. DINOv3 shows consistent performance improvements with increased capacity across all tasks, with DINOv3-Large single-task achieving 100\% classification, 62.10\% detection, and 73.53\% segmentation. ConvNeXt exhibits more complex scaling behavior, with ConvNeXt-Base single-task showing superior detection performance (60.40\%) compared to both smaller and larger variants, suggesting potential optimization challenges at larger scales.

\paragraph{Analysis and Insights}
The experimental results reveal several important characteristics of multi-task learning in automotive anomaly detection:
\begin{itemize}
    \item \textbf{Architecture-Specific MTL Behavior}: The effectiveness of multi-task learning strategies varies significantly between transformer-based (DINOv3)\cite{simeoni2025dinov3} and CNN-based (ConvNeXt)\cite{liu2022convnet} architectures, with ConvNeXt\cite{liu2022convnet} generally showing better adaptability to multi-task coordination under our adaptive strategy.
    
    \item \textbf{Task Interdependence Complexity}: The inconsistent performance patterns across tasks and architectures suggest complex task relationships that cannot be captured by simple weighting schemes. The adaptive strategy shows promise in navigating these complex interdependencies, particularly in CNN architectures.
    
    \item \textbf{Practical Deployment Considerations}: Despite the performance variations, multi-task learning maintains compelling practical advantages. The small performance gaps in many configurations (often <1\%) must be weighed against the significant efficiency benefits of unified models for real-world automotive inspection systems.
    
    \item \textbf{Future Optimization Directions}: The mixed results highlight the need for more sophisticated task coordination mechanisms that can better account for architecture-specific characteristics and complex task relationships in automotive anomaly detection.
\end{itemize}

\subsection{CAD 100K Dataset Validation}

\paragraph{Real-world Performance Assessment.}
To validate the practical applicability of our approach and demonstrate the utility of the CAD 100K dataset, we conduct experiments on the real-world portion of the Car Anomaly Detection dataset. This evaluation focuses on three representative tasks:
\begin{itemize}[leftmargin=*,nosep]
    \item Chassis Domain Classification: Identifying anomaly types in undercarriage components
    \item General Domain Detection: Localizing various anomalies across vehicle surfaces  
    \item Appearance Domain Segmentation: Pixel-level anomaly segmentation on exterior surfaces
\end{itemize}
\begin{table}[t]
\centering
\caption{Performance comparison on three types of tasks in the CAD 100K dataset  configurations: ConvNeXt-S with base and adaptive MTL, and DINOv3-Splus with base and adaptive MTL. Results demonstrate the effectiveness of our adaptive strategy across different backbone architectures.}
\resizebox{0.48\textwidth}{!}{
\begin{tabular}{l|cccc}
\toprule
\textbf{Backbone} & \textbf{Method} & \textbf{Cls. Acc (\%)} & \textbf{Det. mAP (\%)} & \textbf{Seg. mIoU (\%)} \\
\midrule
\multirow{2}{*}{DINOv3-Splus}
& Base & \textbf{92.98} & 55.20 & 56.84 \\
& Adaptive & 89.47 & \textbf{59.60} & \textbf{57.19} \\
\midrule
\multirow{2}{*}{ConvNeXt-S}
& Base & \textbf{92.10} & 59.10 & 52.20 \\
& Adaptive & 90.10 & \textbf{60.10} & \textbf{52.77} \\
\bottomrule
\end{tabular}
}
\label{tab:cad_experiments}
\end{table}

As shown in Table~\ref{tab:cad_experiments}, our adaptive MTL approach maintains performance on real-world data. The performance trends mirror those observed on public benchmarks, with adaptive MTL consistently bridging the gap between single-task and naive multi-task approaches.

\paragraph{Dataset Quality Verification.}
The CAD 100K dataset demonstrates comparable task performance to public benchmarks despite the increased complexity of real-world industrial scenarios. The slight performance decrease is expected given the challenging nature of real-world automotive anomaly detection, which includes diverse lighting conditions, occlusions, and manufacturing variations.

The consistent performance across both public benchmarks and the CAD 100K dataset validates the quality and utility of our collected data for supporting multi-task learning in automotive anomaly detection. More extensive experiments covering additional scenarios and domains are provided in the appendix.
\section{Discussion}
 Extensive empirical studies reveal that while multi-task learning presents inherent challenges in task conflict resolution, it offers significant practical advantages in terms of model efficiency and deployment simplicity for real-world automotive manufacturing environments.
\section{Conclusion}

In this paper, we have introduced CAD 100K, the first comprehensive benchmark specifically designed for car related multi-task visual anomaly detection. Through systematic integration of real-world collection, open-source transformation, and synthetic generation, our dataset provides unprecedented coverage across seven vehicle domains and three fundamental tasks—classification, detection, and segmentation. The carefully designed domain–system–part hierarchy enables fine-grained analysis while supporting unified evaluation across diverse automotive inspection scenarios.

The CAD 100K dataset serves as a standardized platform to drive future research in several key directions: developing more sophisticated task coordination strategies that can better exploit the hierarchical structure of automotive inspection tasks, advancing few-shot learning techniques for rare anomaly types, and exploring the transferability of learned representations across different vehicle domains and manufacturing stages. We believe this benchmark will accelerate progress toward more versatile and efficient visual inspection systems for the automotive industry.
{
    \small
    \bibliographystyle{ieeenat_fullname}
    \bibliography{main}

@String(CVPR= {IEEE Conf. Comput. Vis. Pattern Recog.})

@String(CVPR  = {CVPR})

@article{wang2023cardd,
  title={Cardd: A new dataset for vision-based car damage detection},
  author={Wang, Xinkuang and Li, Wenjing and Wu, Zhongcheng},
  journal={IEEE Transactions on Intelligent Transportation Systems},
  volume={24},
  number={7},
  pages={7202--7214},
  year={2023},
  publisher={IEEE}
}

@inproceedings{susutti2024real,
  title={Real-time Car Part Instance Segmentation: the Comparison of the State-of-the-Art},
  author={Susutti, Wittawin and Laoprom, Siwarat and Sutthipanyo, Thanaphit and Vongsaroj, Kanok and Dilokpatpongsa, Pirun and Wattanapornprom, Warin},
  booktitle={2024 28th International Computer Science and Engineering Conference (ICSEC)},
  pages={1--6},
  year={2024},
  organization={IEEE}
}

@inproceedings{baitieva2024supervised,
  title={Supervised anomaly detection for complex industrial images},
  author={Baitieva, Aimira and Hurych, David and Besnier, Victor and Bernard, Olivier},
  booktitle={Proceedings of the IEEE/CVF Conference on Computer Vision and Pattern Recognition},
  pages={17754--17762},
  year={2024}
}

@inproceedings{huynh2023vehide,
  title={VehiDE Dataset: New dataset for Automatic vehicle damage detection in Car insurance},
  author={Huynh, Nhan T and Tran, Nguyen ND and Huynh, Anh T and Hoang, Van-Dung and Nguyen, Hien D},
  booktitle={2023 15th International Conference on Knowledge and Systems Engineering (KSE)},
  pages={1--6},
  year={2023},
  organization={IEEE}
}

@article{mallios2023vehicle,
  title={Vehicle damage severity estimation for insurance operations using in-the-wild mobile images},
  author={Mallios, Dimitrios and Xiaofei, Li and McLaughlin, Niall and Del Rincon, Jesus Martinez and Galbraith, Clare and Garland, Rory},
  journal={IEEE Access},
  volume={11},
  pages={78644--78655},
  year={2023},
  publisher={IEEE}
}

@article{yan2024comprehensive,
  title={A comprehensive survey of deep transfer learning for anomaly detection in industrial time series: Methods, applications, and directions},
  author={Yan, Peng and Abdulkadir, Ahmed and Luley, Paul-Philipp and Rosenthal, Matthias and Schatte, Gerrit A and Grewe, Benjamin F and Stadelmann, Thilo},
  journal={IEEE Access},
  volume={12},
  pages={3768--3789},
  year={2024},
  publisher={IEEE}
}

@inproceedings{mai2006sampled,
  title={Is sampled data sufficient for anomaly detection?},
  author={Mai, Jianning and Chuah, Chen-Nee and Sridharan, Ashwin and Ye, Tao and Zang, Hui},
  booktitle={Proceedings of the 6th ACM SIGCOMM conference on Internet measurement},
  pages={165--176},
  year={2006}
}

@inproceedings{chen2024survey,
  title={A survey on anomaly detection with few-shot learning},
  author={Chen, Junyang and Wang, Changbo and Hong, Yifan and Mi, Rui and Zhang, Liang-Jie and Wu, Yirui and Wang, Huan and Zhou, Yue},
  booktitle={International Conference on Cognitive Computing},
  pages={34--50},
  year={2024},
  organization={Springer}
}

@article{huyan2022aud,
  title={AUD-Net: A unified deep detector for multiple hyperspectral image anomaly detection via relation and few-shot learning},
  author={Huyan, Ning and Zhang, Xiangrong and Quan, Dou and Chanussot, Jocelyn and Jiao, Licheng},
  journal={IEEE Transactions on Neural Networks and Learning Systems},
  volume={35},
  number={5},
  pages={6835--6849},
  year={2022},
  publisher={IEEE}
}

@article{wei2024few,
  title={Few-shot online anomaly detection and segmentation},
  author={Wei, Shenxing and Wei, Xing and Ma, Zhiheng and Dong, Songlin and Zhang, Shaochen and Gong, Yihong},
  journal={Knowledge-Based Systems},
  volume={300},
  pages={112168},
  year={2024},
  publisher={Elsevier}
}

@inproceedings{fang2023fastrecon,
  title={Fastrecon: Few-shot industrial anomaly detection via fast feature reconstruction},
  author={Fang, Zheng and Wang, Xiaoyang and Li, Haocheng and Liu, Jiejie and Hu, Qiugui and Xiao, Jimin},
  booktitle={Proceedings of the IEEE/CVF International Conference on Computer Vision},
  pages={17481--17490},
  year={2023}
}

@article{hernandez1998real,
  title={Real-world data is dirty: Data cleansing and the merge/purge problem},
  author={Hern{\'a}ndez, Mauricio A and Stolfo, Salvatore J},
  journal={Data mining and knowledge discovery},
  volume={2},
  number={1},
  pages={9--37},
  year={1998},
  publisher={Springer}
}

@inproceedings{oquab2014learning,
  title={Learning and transferring mid-level image representations using convolutional neural networks},
  author={Oquab, Maxime and Bottou, Leon and Laptev, Ivan and Sivic, Josef},
  booktitle={Proceedings of the IEEE conference on computer vision and pattern recognition},
  pages={1717--1724},
  year={2014}
}

@article{wei2021review,
  title={A review on evolutionary multitask optimization: Trends and challenges},
  author={Wei, Tingyang and Wang, Shibin and Zhong, Jinghui and Liu, Dong and Zhang, Jun},
  journal={IEEE Transactions on Evolutionary Computation},
  volume={26},
  number={5},
  pages={941--960},
  year={2021},
  publisher={IEEE}
}

@article{berwo2024vebd,
  title={VEBD-HEL: A noval approach to vehicle exterior body damage parts classification in intelligent transportation systems},
  author={Berwo, Michael Abebe and Fang, Yong and Khan, Asad and Manzor, Adnan and Mahmood, Jabar},
  journal={Alexandria Engineering Journal},
  volume={108},
  pages={961--975},
  year={2024},
  publisher={Elsevier}
}

@article{omar2013machine,
  title={Machine learning techniques for anomaly detection: an overview},
  author={Omar, Salima and Ngadi, Asri and Jebur, Hamid H},
  journal={International Journal of Computer Applications},
  volume={79},
  number={2},
  year={2013},
  publisher={Foundation of Computer Science}
}

@inproceedings{song2023objectstitch,
  title={Objectstitch: Object compositing with diffusion model},
  author={Song, Yizhi and Zhang, Zhifei and Lin, Zhe and Cohen, Scott and Price, Brian and Zhang, Jianming and Kim, Soo Ye and Aliaga, Daniel},
  booktitle={Proceedings of the IEEE/CVF Conference on Computer Vision and Pattern Recognition},
  pages={18310--18319},
  year={2023}
}

@inproceedings{dai2025seas,
  title={SeaS: few-shot industrial anomaly image generation with separation and sharing fine-tuning},
  author={Dai, Zhewei and Zeng, Shilei and Liu, Haotian and Li, Xurui and Xue, Feng and Zhou, Yu},
  booktitle={Proceedings of the IEEE/CVF International Conference on Computer Vision},
  pages={23135--23144},
  year={2025}
}

@inproceedings{yun2023achievement,
  title={Achievement-based training progress balancing for multi-task learning},
  author={Yun, Hayoung and Cho, Hanjoo},
  booktitle={Proceedings of the IEEE/CVF International Conference on Computer Vision},
  pages={16935--16944},
  year={2023}
}

@article{hasan2025vehicle,
  title={Vehicle Damage Detection Using Artificial Intelligence: A Systematic Literature Review},
  author={Hasan, Md Jahid and Nguyen, Cong Kha and Boo, Yee Ling and Jahani, Hamed and Ong, Kok-Leong},
  journal={Wiley Interdisciplinary Reviews: Data Mining and Knowledge Discovery},
  volume={15},
  number={2},
  pages={e70027},
  year={2025},
  publisher={Wiley Online Library}
}

@article{peng2025car,
  title={Car Damage Detection Based on Multi-View Fusion and Alignment: Dataset and Method},
  author={Peng, Jinbo and Dong, Shoubin and Yuan, Hua and Zheng, Xiaorou},
  journal={IEEE Transactions on Intelligent Transportation Systems},
  year={2025},
  publisher={IEEE}
}

@article{graham2023one,
  title={One model is all you need: multi-task learning enables simultaneous histology image segmentation and classification},
  author={Graham, Simon and Vu, Quoc Dang and Jahanifar, Mostafa and Raza, Shan E Ahmed and Minhas, Fayyaz and Snead, David and Rajpoot, Nasir},
  journal={Medical Image Analysis},
  volume={83},
  pages={102685},
  year={2023},
  publisher={Elsevier}
}

@article{simeoni2025dinov3,
  title={Dinov3},
  author={Sim{\'e}oni, Oriane and Vo, Huy V and Seitzer, Maximilian and Baldassarre, Federico and Oquab, Maxime and Jose, Cijo and Khalidov, Vasil and Szafraniec, Marc and Yi, Seungeun and Ramamonjisoa, Micha{\"e}l and others},
  journal={arXiv preprint arXiv:2508.10104},
  year={2025}
}

@article{lin2022augmentation,
  title={Augmentation dataset of a two-dimensional neural network model for use in the car parts segmentation and car classification of three dimensions},
  author={Lin, Chuen-Horng and Yu, Chia-Ching and Chen, Huan-Yu},
  journal={The Journal of Supercomputing},
  volume={78},
  number={17},
  pages={18915--18958},
  year={2022},
  publisher={Springer}
}

@inproceedings{liu2022convnet,
  title={A convnet for the 2020s},
  author={Liu, Zhuang and Mao, Hanzi and Wu, Chao-Yuan and Feichtenhofer, Christoph and Darrell, Trevor and Xie, Saining},
  booktitle={Proceedings of the IEEE/CVF conference on computer vision and pattern recognition},
  pages={11976--11986},
  year={2022}
}

@article{zhang2024pku,
  title={PKU-GoodsAD: A supermarket goods dataset for unsupervised anomaly detection and segmentation},
  author={Zhang, Jian and Ding, Runwei and Ban, Miaoju and Dai, Linhui},
  journal={IEEE Robotics and Automation Letters},
  year={2024},
  publisher={IEEE}
}

@article{li2024co,
  title={Co-training transformer for remote sensing image classification, segmentation, and detection},
  author={Li, Qingyun and Chen, Yushi and He, Xin and Huang, Lingbo},
  journal={IEEE Transactions on Geoscience and Remote Sensing},
  volume={62},
  pages={1--18},
  year={2024},
  publisher={IEEE}
}

@article{yang2025cabs,
  title={CABS: Conflict-Aware and Balanced Sparsification for Enhancing Model Merging},
  author={Yang, Zongzhen and Qi, Binhang and Sun, Hailong and Long, Wenrui and Zhao, Ruobing and Gao, Xiang},
  journal={arXiv preprint arXiv:2503.01874},
  year={2025}
}

@inproceedings{chen2018gradnorm,
  title={Gradnorm: Gradient normalization for adaptive loss balancing in deep multitask networks},
  author={Chen, Zhao and Badrinarayanan, Vijay and Lee, Chen-Yu and Rabinovich, Andrew},
  booktitle={International conference on machine learning},
  pages={794--803},
  year={2018},
  organization={PMLR}
}

@article{cai2023itran,
  title={ITran: A novel transformer-based approach for industrial anomaly detection and localization},
  author={Cai, Xiangyu and Xiao, Ruliang and Zeng, Zhixia and Gong, Ping and Ni, Youcong},
  journal={Engineering Applications of Artificial Intelligence},
  volume={125},
  pages={106677},
  year={2023},
  publisher={Elsevier}
}

@inproceedings{mvtec_ad,
  title={MVTec AD--A comprehensive real-world dataset for unsupervised anomaly detection},
  author={Bergmann, Paul and Fauser, Michael and Sattlegger, David and Steger, Carsten},
  booktitle={CVPR},
  pages={9592--9600},
  year={2019}
}

@inproceedings{real_ad,
  title={Real-iad: A real-world multi-view dataset for benchmarking versatile industrial anomaly detection},
  author={Wang, Chengjie and Zhu, Wenbing and Gao, Bin-Bin and Gan, Zhenye and Zhang, Jiangning and Gu, Zhihao and Qian, Shuguang and Chen, Mingang and Ma, Lizhuang},
  booktitle={CVPR},
  pages={22883--22892},
  year={2024}
}

@article{schmidl2022anomaly,
  title={Anomaly detection in time series: a comprehensive evaluation},
  author={Schmidl, Sebastian and Wenig, Phillip and Papenbrock, Thorsten},
  journal={Proceedings of the VLDB Endowment},
  volume={15},
  number={9},
  pages={1779--1797},
  year={2022},
  publisher={VLDB Endowment}
}

@inproceedings{deng2022anomaly,
  title={Anomaly detection via reverse distillation from one-class embedding},
  author={Deng, Hanqiu and Li, Xingyu},
  booktitle={CVPR},
  pages={9737--9746},
  year={2022}
}

@article{gong2024coba,
  title={Coba: convergence balancer for multitask finetuning of large language models},
  author={Gong, Zi and Yu, Hang and Liao, Cong and Liu, Bingchang and Chen, Chaoyu and Li, Jianguo},
  journal={arXiv preprint arXiv:2410.06741},
  year={2024}
}
}
\clearpage
\setcounter{page}{1}
\renewcommand\thefigure{B\arabic{figure}}
\renewcommand\thetable{B\arabic{table}}  
\renewcommand\theequation{B\arabic{equation}}
\setcounter{equation}{0}
\setcounter{table}{0}
\setcounter{figure}{0}\renewcommand\thefigure{B\arabic{figure}}
\renewcommand\thetable{B\arabic{table}}  
\renewcommand\theequation{B\arabic{equation}}
\appendix
\section{Ethical Considerations and Data Privacy}
\label{sec:ethics}

In accordance with the ethics guidelines, we have implemented comprehensive measures to protect privacy and mitigate potential negative societal impacts throughout the creation of the CAD 100K dataset.

\subsection{Privacy Protection Measures}
During the data collection and processing phases, we systematically removed all personally identifiable information and sensitive data:

\begin{itemize}
    \item \textbf{License Plates}: All vehicle license plates were automatically detected and blurred using computer vision algorithms, followed by manual verification to ensure complete anonymization.
    
    \item \item \textbf{Vehicle Logos and Identifiers}: Manufacturer logos, vehicle model badges, and other identifying marks were removed or obscured to prevent brand identification and commercial sensitivity concerns.
    
    \item \textbf{Human Subjects}: Any images containing human faces, body parts, or other personally identifiable human characteristics were either excluded from the dataset or underwent rigorous anonymization processing.
    
    \item \textbf{Location Context}: Background elements that could reveal specific geographical locations or private property details were carefully processed to maintain privacy.
\end{itemize}

\section{ Dataset Visualization and Supplementary Experiments}
\label{sec:rationale}

\begin{figure*}[t]
    \centering
    \includegraphics[width=1\linewidth]{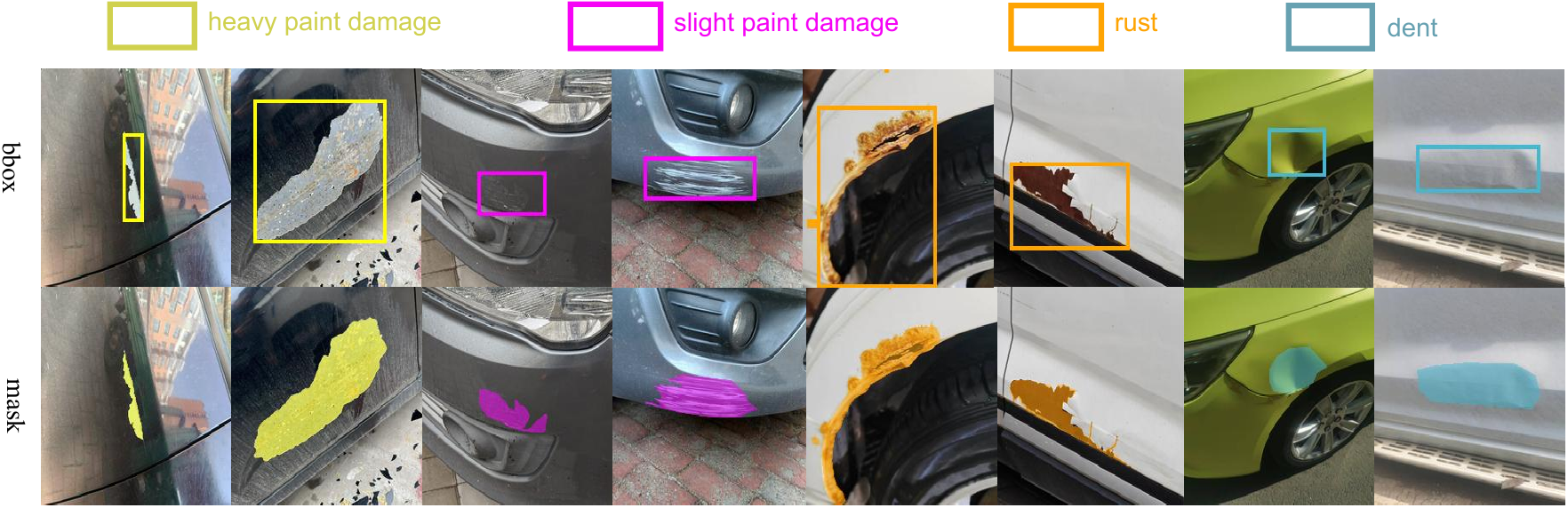}
     \caption{\textbf{The Exterior Domain Subset Supporting both Detection and Segmentation in the CAD 100K Dataset.} This subset contains various exterior anomalies with both bounding box and pixel-level mask annotations.}
    \label{fig:sub1}
\end{figure*}

\begin{figure*}[t]
    \centering
    \includegraphics[width=1\linewidth]{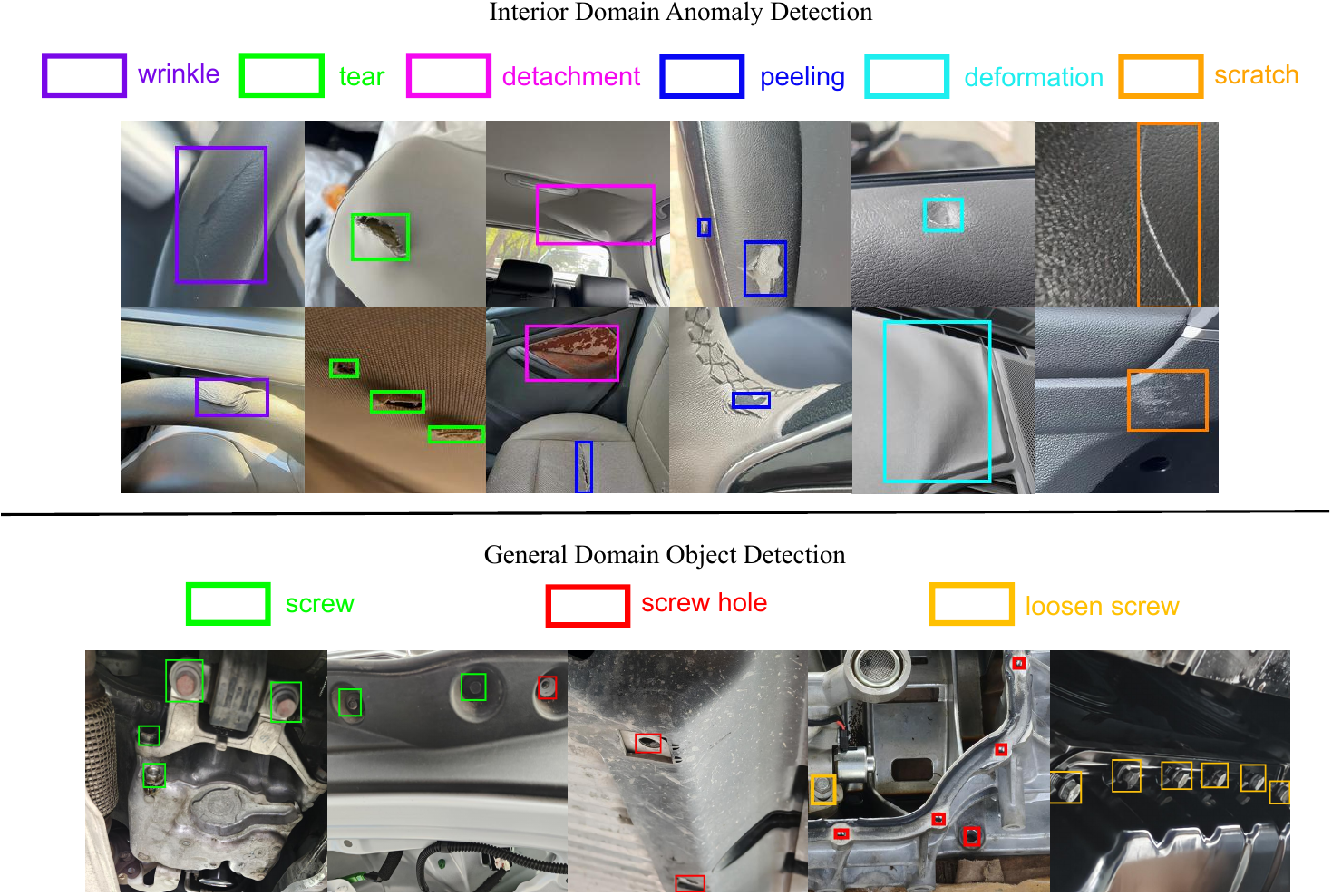}
     \caption{\textbf{The Interior and General Domain Detection Subset in the CAD 100K Dataset.} Examples include interior component defects and general cross-system anomalies with detection annotations.}
    \label{fig:sub2}
\end{figure*}

This appendix provides visual documentation of the CAD 100K dataset subsets, supporting the experimental validation in the main paper. Figure \ref{fig:sub1} illustrates the exterior domain subset that simultaneously supports both detection and segmentation tasks, featuring various anomaly types with corresponding bounding boxes and segmentation masks. Figure \ref{fig:sub2} showcases the interior and general domain detection subsets, highlighting diverse anomaly patterns in cabin components and cross-system defects.

\begin{figure*}[t]
    \centering
    \includegraphics[width=0.75\linewidth]{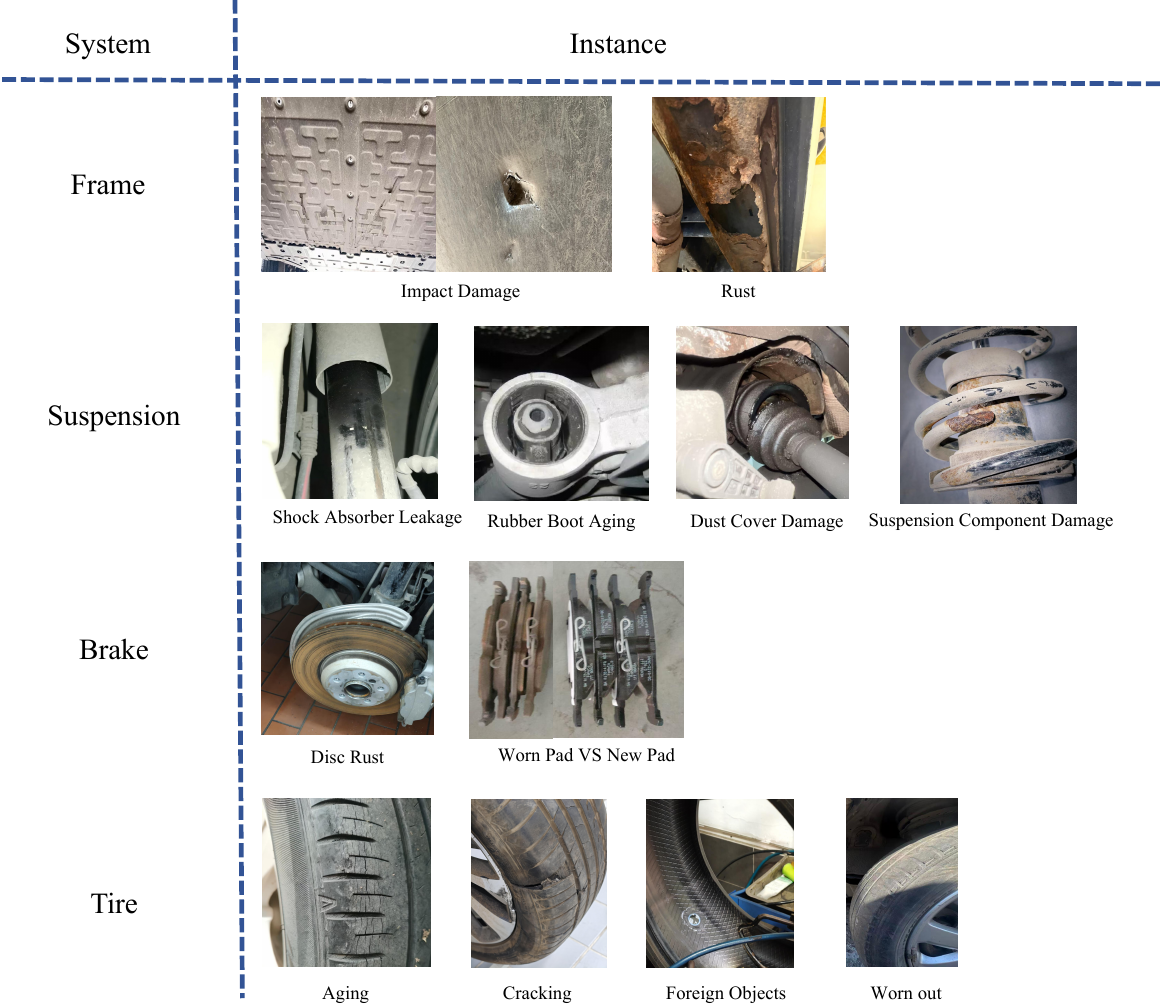}
     \caption{\textbf{The Chassis Domain Classification Subset in the CAD 100K Dataset.} Covering suspension, brake, and tire systems with various anomaly types.}
    \label{fig:sub3}
\end{figure*}

\begin{figure*}[t]
    \centering
    \includegraphics[width=0.75\linewidth]{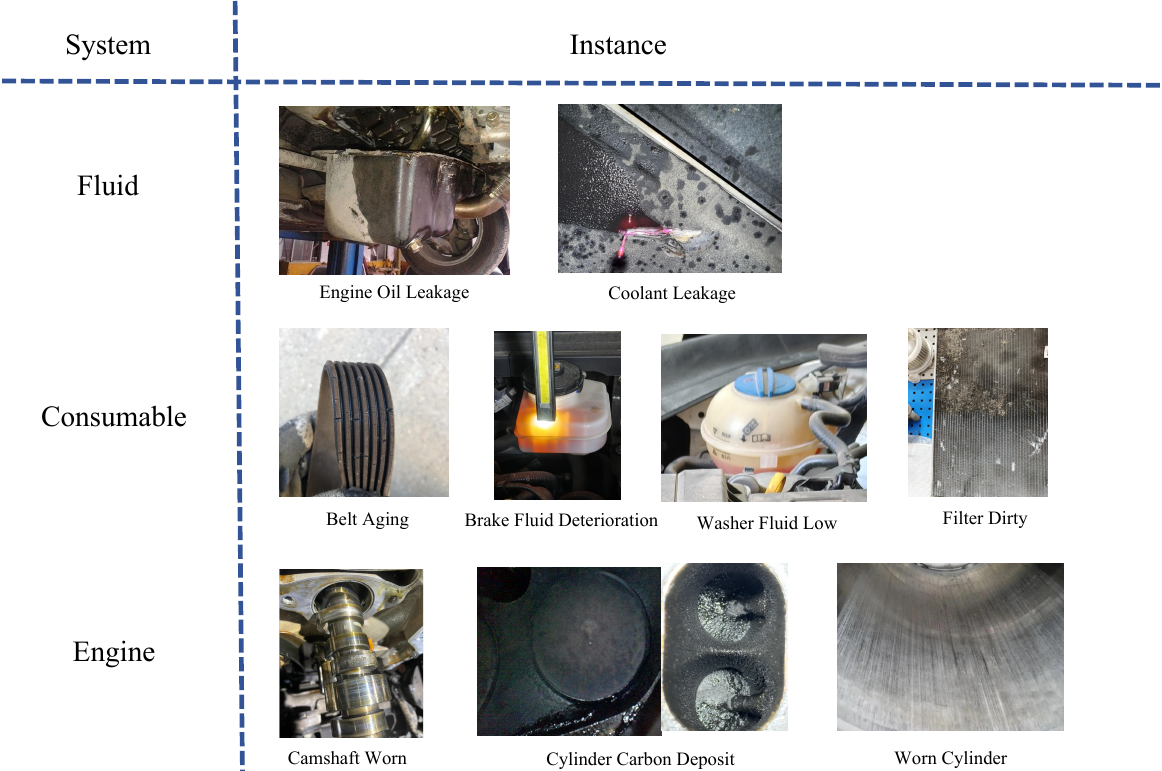}
     \caption{\textbf{The Engine Domain Classification Subset in the CAD 100K Dataset.} Featuring engine components, fluid systems, and mechanical parts.}
    \label{fig:sub4}
\end{figure*}

\begin{figure*}[t]
    \centering
    \includegraphics[width=0.75\linewidth]{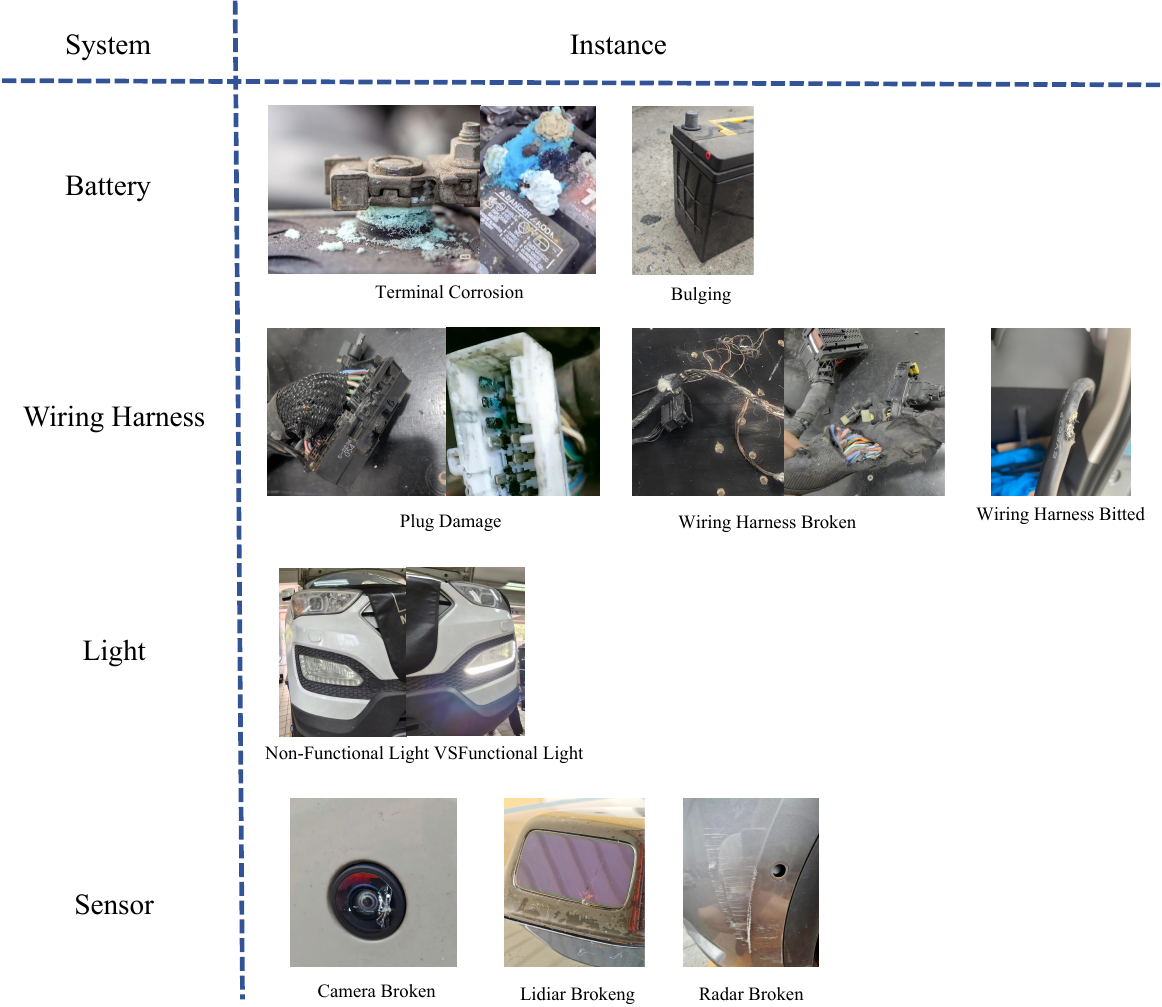}
     \caption{\textbf{The Electrical Domain Classification Subset in the CAD 100K Dataset.} Including battery systems, wiring harnesses, and electronic components.}
    \label{fig:sub5}
\end{figure*}

\begin{figure*}[t]
    \centering
    \includegraphics[width=0.75\linewidth]{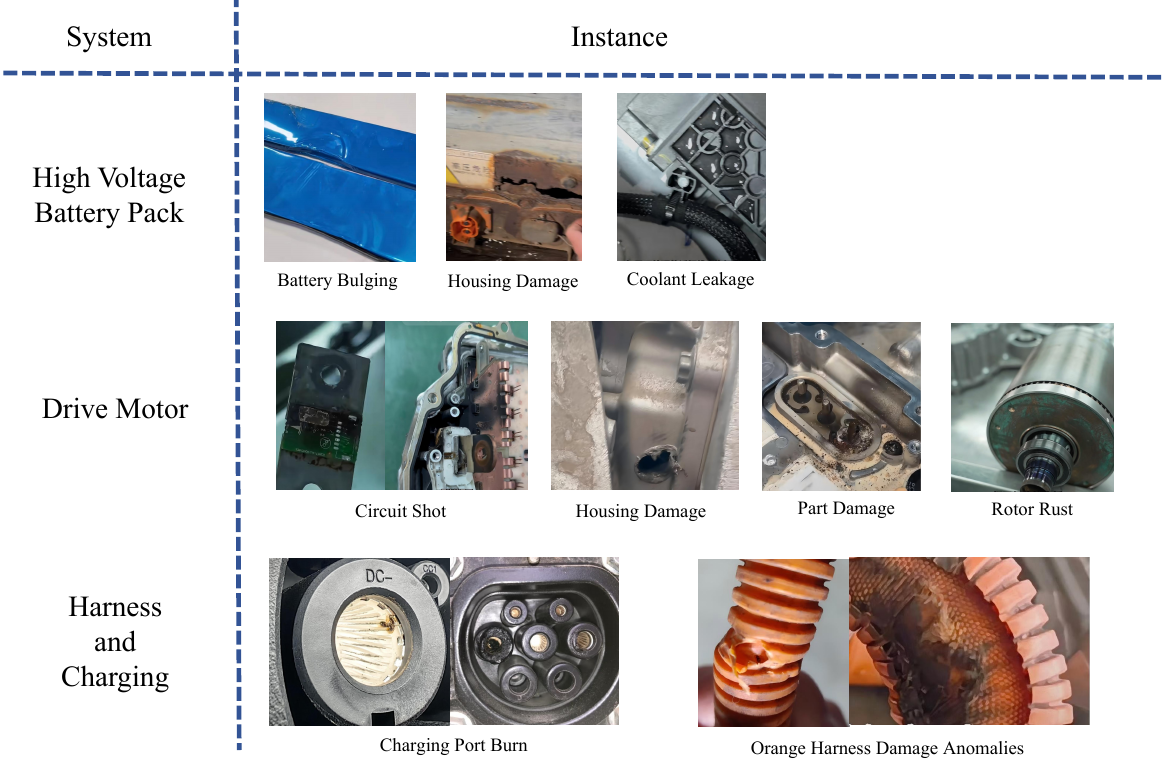}
     \caption{\textbf{The EV-specific Domain Classification Subset in the CAD 100K Dataset.} Specialized components for electric vehicles including high-voltage systems and charging infrastructure.}
    \label{fig:sub6}
\end{figure*}

Figures \ref{fig:sub3} through \ref{fig:sub6} present the classification subsets across four specialized vehicle component domains: chassis systems (suspension, brakes, tires), engine bay components (fluids, consumables, mechanical parts), electrical systems (battery, wiring, sensors), and EV-specific components (high-voltage systems, drive motors, charging infrastructure). These visual examples demonstrate the dataset's comprehensive coverage of automotive anomalies across different vehicle systems and domains.

The hierarchical organization and multi-task annotations visible in these figures enable the rigorous evaluation of multi-task learning approaches presented in our experimental results, providing a solid foundation for advancing automotive visual inspection technologies.

Besides, we conducted additional experiments to further analyze the multi-task learning performance across different backbone architectures and training strategies. Table \ref{tab:extended_mtl_analysis} provides a comprehensive comparison of computation efficiency and parameter utilization.

\begin{table*}[t]
\caption{\textbf{Detection Performance Comparison Across Vehicle Domains}. Models are evaluated on three vehicle domains (Interior, General, Exterior) using standard COCO metrics including AP at different IoU thresholds and AR.}
\centering
\begin{tabular}{lcccc|cccc|cccc}
\toprule
\multirow{2}{*}{Model} & \multicolumn{4}{c}{Interior} & \multicolumn{4}{c}{General} & \multicolumn{4}{c}{Exterior} \\
\cmidrule(lr){2-5} \cmidrule(lr){6-9} \cmidrule(lr){10-13}
 & AP & AP$_{50}$ & AP$_{75}$ & AR & AP & AP$_{50}$ & AP$_{75}$ & AR & AP & AP$_{50}$ & AP$_{75}$ & AR \\
\midrule
DINOv3-Splus & \textbf{48.0} & \textbf{79.9} & \textbf{46.7} & 61.0 & 62.4 & 91.2 & 65.88 & 68.9 & 45.1 & 66.9 & 45.2 & 54.5 \\
ConvNeXt-S   & 46.9 & 76.1 & 46.6 & \textbf{61.8} & \textbf{64.8} & \textbf{92.8} & \textbf{72.3} & \textbf{72.4} & \textbf{51.0} & \textbf{72.0} & \textbf{54.4} & \textbf{57.6} \\
\bottomrule
\end{tabular}
\label{tab:detection_results}
\end{table*}

\begin{table*}[t]
\caption{\textbf{Classification Performance Across Vehicle Component Domains}. Models are evaluated on four vehicle component domains using top-1 and top-5 accuracy metrics.}
\centering
\small
\begin{tabular}{lcc|cc|cc|cc}
\toprule
\multirow{2}{*}{Model} & \multicolumn{2}{c}{Chassis} & \multicolumn{2}{c}{Engine} & \multicolumn{2}{c}{Electric} & \multicolumn{2}{c}{EV-specific} \\
\cmidrule(lr){2-3} \cmidrule(lr){4-5} \cmidrule(lr){6-7} \cmidrule(lr){8-9}
 & Acc@1 & Acc@5 & Acc@1 & Acc@5 & Acc@1 & Acc@5 & Acc@1 & Acc@5 \\
\midrule
DINOv3-Splus & \textbf{91.2} & 100.0 & 90.1 & 100.0 & 93.5 & 99.1 & 97.9 & 100.0 \\
ConvNeXt-S & 90.4 & 100.0 & \textbf{93.4} & 100.0 & \textbf{93.5} & \textbf{100.0} & \textbf{100.0} & 100.0 \\
\bottomrule
\end{tabular}
\label{tab:classification_results}
\end{table*}

\begin{table*}[t]
\caption{\textbf{Segmentation Performance Comparison}. Models are evaluated on both car exterior anomalies segmentation and car parts segmentation tasks using mIoU, mFscore, and mPrecision metrics.}
\centering
\small
\begin{tabular}{lcccc|cccc}
\toprule
\multirow{2}{*}{Model} & \multicolumn{4}{c}{Car Exterior Anomalies} & \multicolumn{4}{c}{Car Parts Segmentation} \\
\cmidrule(lr){2-5} \cmidrule(lr){6-9}
 & mIoU & mFscore & mPrec & mRecall & mIoU & mFscore & mPrec & mRecall \\ 
\midrule
DINOv3-Splus & \textbf{54.9} & \textbf{67.0} & 85.7 & \textbf{58.1} & 72.0 & 81.7 & 81.9 & 81.9 \\
ConvNeXt-S & 52.4 & 65.1 & \textbf{90.1} & 54.4 & \textbf{72.9} & \textbf{82.4} & \textbf{82.7} & \textbf{82.6} \\
\bottomrule
\end{tabular}
\label{tab:segmentation_results}
\end{table*}

\begin{table*}[t]
\caption{\textbf{Multi-task Learning Performance Comparison on Public Benchmarks}. Models are evaluated across classification, detection, and segmentation tasks using single-task learning (STL), base multi-task learning (Base-MTL), and adaptive multi-task learning (Adaptive-MTL) approaches.}
\centering
\begin{tabular}{lcccc}
\toprule
Model & Method & Cls. Acc (\%) & Det. mAP (\%) & Seg. mIoU (\%) \\
\midrule
\multirow{3}{*}{DINOv3-Splus}
& Single & 98.40 & \textbf{59.20} & \textbf{72.80} \\
& Base-MTL & 98.40 & 58.90 & 70.93 \\
& Adaptive-MTL & 98.40 & 57.80 & 71.88 \\
\midrule
\multirow{3}{*}{ConvNeXt-S}
& Single & \textbf{98.80} & 58.80 & \textbf{72.80} \\
& Base-MTL & \textbf{98.80} & 58.60 & 71.12 \\
& Adaptive-MTL & \textbf{98.80} & 59.10 & 71.82 \\
\midrule
YOLOv11s-cls & Single & 97.2 & - & - \\
YOLOv11s-det & Single & - & 58.0 & - \\
ResNet-101+PSPNet & Single & - & - & 63.9 \\
\bottomrule
\end{tabular}
\label{tab:multitask_comparison}
\end{table*}

\end{document}